\documentclass[letterpaper, 10 pt, conference]{ieeeconf}  

\IEEEoverridecommandlockouts                              

\usepackage{graphics} 
\usepackage{epsfig} 
\usepackage{mathptmx} 
\usepackage{times} 
\usepackage{amsmath} 
\usepackage{amssymb}  
\usepackage{booktabs}
\usepackage{adjustbox}
\usepackage{bm}
\usepackage{graphicx}
\usepackage{rotating}
\usepackage{dblfloatfix}
\usepackage[ruled,vlined,noend]{algorithm2e}
\usepackage{hyperref}
\usepackage[usenames,dvipsnames]{xcolor} 

\title{\LARGE \bf Measure Anything: Real-time, Multi-stage Vision-based Dimensional Measurement using Segment Anything
}

\author{Yongkyu Lee$^{1}$, Shivam K Panda$^{1}$, Wei Wang$^{2}$ and Mohammad Khalid Jawed$^{1}$
\thanks{This work was supported by the United States Department of Agriculture (USDA Award No. 2024-67021-42528 \& 2022-67022-37021).}
\thanks{$^{1}$Yongkyu Lee, Shivam K Panda and Mohammad Khalid Jawed are with the Department of Mechanical and Aerospace Engineering, University of California, Los Angeles (UCLA), Los Angeles, CA 90095, USA
        ({\tt\small yongkyulee@g.ucla.edu; shivamkp@g.ucla.edu; khalidjm@seas.ucla.edu})}%
\thanks{$^{2}$Wei Wang is with the Department of Computer Science, University of California, Los Angeles (UCLA), Los Angeles, CA 90095, USA
        ({\tt\small weiwang@cs.ucla.edu})}%
}

\begin{document}

\maketitle
\thispagestyle{empty}
\pagestyle{empty}

\begin{abstract}

We present Measure Anything, a comprehensive vision-based framework for dimensional measurement of objects with circular cross-sections, leveraging the Segment Anything Model (SAM). Our approach estimates key geometric features—including diameter, length, and volume—for rod-like geometries with varying curvature and general objects with constant skeleton slope. The framework integrates segmentation, mask processing, skeleton construction, and 2D-3D transformation, packaged in a user-friendly interface. We validate our framework by estimating the diameters of Canola stems -- collected from agricultural fields in North Dakota -- which are thin and non-uniform, posing challenges for existing methods. Measuring its diameters is critical, as it is a phenotypic traits that correlates with the health and yield of Canola crops. This application also exemplifies the potential of Measure Anything, where integrating intelligent models -- such as keypoint detection -- extends its scalability to fully automate the measurement process for high-throughput applications. Furthermore, we showcase its versatility in robotic grasping, leveraging extracted geometric features to identify optimal grasp points. Our dataset and code repository are available at \href{https://github.com/StructuresComp/measure-anything}{https://github.com/StructuresComp/measure-anything}
\end{abstract}

 \begin{figure*}[!t]
    \centering
     \includegraphics[width=\textwidth]{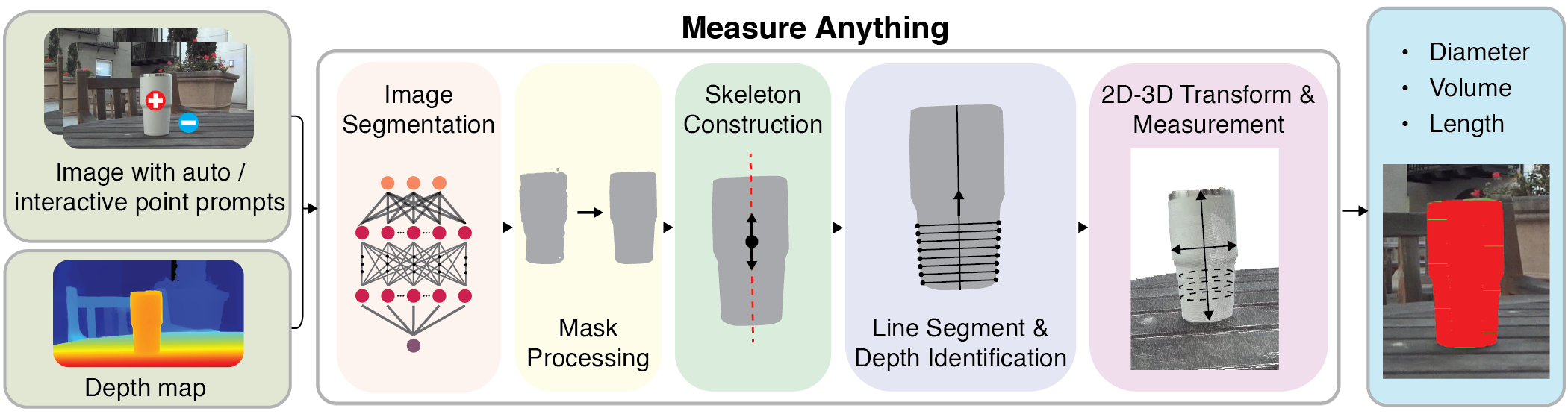}
    \caption{Overview of the Measure Anything framework}
    \label{fig:framework_overview}
\end{figure*}

\section{INTRODUCTION}\label{sec:intro}
Advances in robotic perception empower machines to better understand and interact with their environments, enabling them to perform complex, labor-intensive tasks with greater consistency, efficiency, and scalability. Among these tasks, vision-based dimensional measurement plays a critical role across diverse domains: in precision agriculture, accurate measurement of plant dimensions optimizes resource allocation and enhances crop yields; in manufacturing, it ensures product quality through precise dimensional verification, reducing waste and defects; and in robotic manipulation, it enables robust object interactions in tasks such as assembly, packaging, and material handling. Despite its widespread significance, vision-based dimensional measurement faces key challenges, including identifying target objects in cluttered environments filled with complex geometries and occlusions, and implementing systematic methods to extract accurate measurements from object contours. 

Measure Anything addresses the first challenge by integrating the Segment Anything Model (SAM) \cite{kirillov2023segment, ravi2024sam}, a state-of-the-art segmentation foundation model with exceptional zero-shot generalization capabilities. Measure Anything inherits the promptable features of SAM and explores two approaches to obtaining point prompts that guide object selection. The first is an interactive method, where users place positive or negative point prompts directly on the object of interest. The second is an automated method, where a keypoint detection model replaces manual input, enabling high scalability and efficient operation in large-scale scenarios. Beyond segmentation, Measure Anything employs a systematic pipeline to extract dimensional measurements from object masks. This pipeline includes sequential tasks such as mask processing, skeleton construction, line segment identification, and 2D-3D transformation. Tailored skeleton construction methods are applied based on the object’s geometry, enabling the accurate computation of diameter, length, and volume in a single pass. An overview of Measure Anything framework can be seen in Figure \ref{fig:framework_overview}. By combining cutting-edge segmentation capabilities with a robust geometric processing, Measure Anything offers a comprehensive solution to the challenges of vision-based dimensional measurement.

This paper focuses on two primary applications: precision agriculture and robotic grasping. For precision agriculture, we tackle the task of estimating stem diameters, a key phenotypic trait that provides valuable insights into crop health, environmental responses, genetic factors, and overall yield potential \cite{de2015stem}. For robotic grasping, we show that the dimensional measurements obtained from Measure Anything can be applied to general manipulation tasks such as identifying optimal grip points. Both applications involve target objects placed in highly cluttered environments, presenting significant challenges for accurate perception. Beyond these primary use cases, the framework’s modular design allows for extensions to other applications, such as quality inspection in manufacturing or crack-width measurement in structural assessments, with minimal modifications.

In summary, the key contributions of this work are as follows:
\begin{enumerate}
\item Development of a robust multi-stage pipeline integrating SAM, and downstream operations for accurately measuring the diameter, length, and volume of target objects with circular cross-sections.
\item Implementation of the pipeline through a user-friendly demo that can be used on 
various objects with minimal modifications
\item Validation of the pipeline's effectiveness on measuring Canola stem diameters, as well diameters and demonstration of its scalability when combined with a keypoint detection model for automation.
\item Application to general manipulation tasks, where the framework's extracted geometric features are utilized in a stability model to determine optimal grasping coordinates.
\end{enumerate}

Our paper is organized as follows. Section \ref{sec:related_works} contains a literature review of various topics related to our framework. Section \ref{sec:methodology} provides a detailed explanation of each stage of our pipeline, and the results are presented in Section \ref{sec:results}. 

\section{RELATED WORKS}\label{sec:related_works}

Various sensors all used for dimension measurement tasks across various applications, with their popularity depending on factors such as precision, scalability, and repeatability. Commonly used sensors include LiDAR \cite{kim2024p}, cameras \cite{zhou2021rgb}, 3D scanners \cite{wang2021development}, and photogrammetry setups \cite{shang2022measurement}. Each of these sensors offers distinct advantages: LiDAR excels at high-precision distance measurements and performs reliably in diverse environmental conditions; 3D scanners capture detailed surface geometries with high accuracy; and photogrammetry frameworks reconstruct three-dimensional structures from two-dimensional images, enabling flexible and scalable setups. Among these, vision-based techniques using stereo or depth cameras have emerged as particularly popular due to their combination of accuracy, ease of deployment, and scalability. These methods are particularly advantageous for measurements in cluttered or remote environments, where other sensors may face limitations. This research focuses on applications where vision-based measurement techniques offer the most feasible and effective solutions.

In precision agriculture, vision-based methods are extensively used to estimate fruit sizes, which directly influence harvesting decisions \cite{kang2020fruit, mirbod2023tree}. Other physical traits such as stem diameter, leaf length, plant height, branch angle, are critical for crop phenotyping, crop monitoring and yield forecasting tasks. For example, Vit \textit{et al.} \cite{vit2019length} proposed a length-based phenotyping technique applicable to various plants, such as banana leaves and cucumbers. Their method involves three stages: object detection, point of interest identification, and 3D measurement. Similarly, a high-throughput stereo vision system called “PhenoStereo” \cite{xiang2021measuring} was developed for in-field stem diameter measurement of sorghum plants. By leveraging Mask R-CNN-based instance segmentation \cite{he2017mask} and semi-global block matching (SGBM) \cite{hirschmuller2007stereo} for 3D point cloud reconstruction, PhenoStereo achieved high accuracy and strong correlation with ground truth measurements. Similarly, in livestock farming, vision-based measurement techniques automate the assessment of livestock body dimensions \cite{ma2024computer}, playing a vital role in breeding management and the selection of superior livestock breeds \cite{fernandes2020image}.

In robotic manipulation, grasping objects of varying shapes and sizes is a well-researched area \cite{bicchi2000robotic, marwan2021comprehensive, zhang2020state}. Vision-based frameworks play a critical role in estimating grasp points and performing pose estimation of grippers \cite{du2021vision}. These operations rely on accurate measurement of object dimensions, assessment of their shapes, and identification of optimal contact points for stable grasping \cite{cheng2022vision}.
Fang \textit{et al.} \cite{fang2020graspnet} devised Graspnet, an end-to-end framework that directly generates abundant grasp poses for an input scene point cloud. Wang \textit{et al.} \cite{wang2021graspness} proposed ``graspness", a geometry-based metric that distinguishes graspable areas in cluttered scenes, significantly improving grasp success rates. Building on this concept, Fang \textit{et al.} \cite{fang2023anygrasp} incorporated center-of-gravity awareness into grasp perception, enabling fast and accurate continuous grasp pose detection with a parallel gripper. This advancement further enhanced the stability and reliability of robotic grasping in dynamic and cluttered environments.

\begin{figure*}[h]
\centering
     \includegraphics[width=\textwidth]{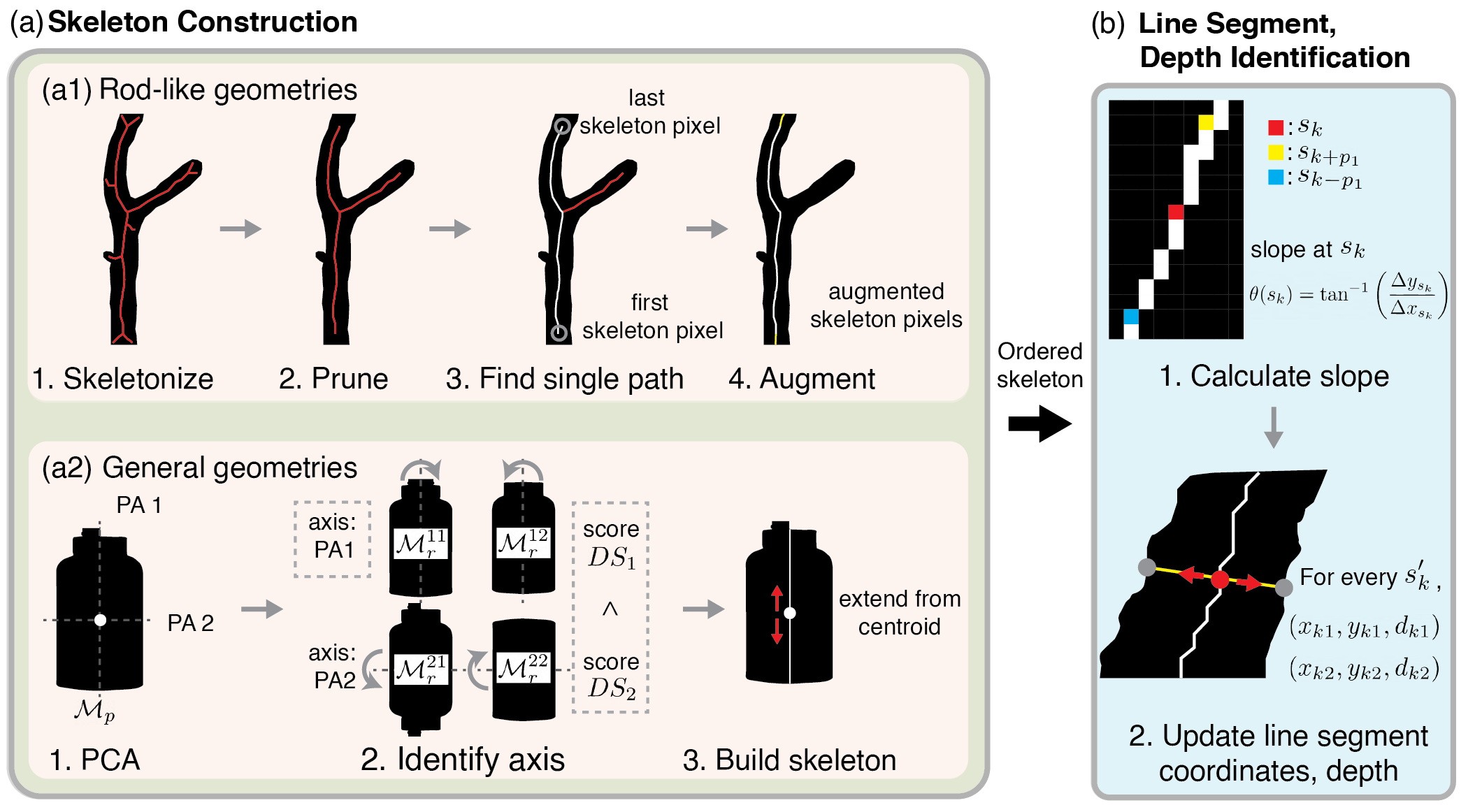}
     \caption{Skeleton Construction and Line Segment Depth Identification Modules. (a) Workflow for skeleton construction modules tailored to rod-like and general geometries. (b) Steps for line segment and depth identification modules.}
     \label{fig:skeletonization}
\end{figure*}

Image segmentation is a fundamental task in computer vision, aiming to classify each pixel of an image into a semantic label \cite{minaee2021image}. In many of the aforementioned applications, semantic segmentation serves as the initial step to isolate regions of interest. Over the past decade, Convolutional Neural Network (CNN)-based models such as Fully Convolution Networks (FCNs) \cite{long2015fully}, encoder-decoder based models \cite{noh2015learning, badrinarayanan2017segnet}, and U-Nets \cite{ronneberger2015u, cciccek20163d}  have been widely adopted. More recently, transformer-based models, such as Vision Transformers (ViT) \cite{dosovitskiy2020image} and Detection Transformers (DETR) \cite{carion2020end, li2023mask} have gained prominence. Due to their superior global context modeling, these models have achieved state-of-the-art performance in numerous segmentation tasks. The latest breakthrough comes from promptable foundation models such as Segment Anything Models (SAM 1 \cite{kirillov2023segment} and SAM 2 \cite{ravi2024sam}). These models feature a hybrid architecture that integrates CNN-based encoders with transformer-based prompt encoders and mask decoders. SAM's remarkable zero-shot generalizability enables effective segmentation across diverse tasks without requiring additional data, positioning these models as game changers in the field.

\section{METHODOLOGY}\label{sec:methodology}

\subsection{Promptable Image Segmentation with SAM 2}\label{sec:segmentation}

The pipeline begins with the identification of the object of interest through a binary mask. To achieve this, we leverage SAM 2 for its exceptional zero-shot generalization capabilities, which allow accurate segmentation without the need a tailored segmentation model trained on custom data. The segmentation process is guided by point prompts, which can be either positive or negative. Positive prompts guide the model to identify masks including itself, while negative prompts resolve ambiguities by excluding unwanted areas, particularly in cases where hierarchical overlap exists.

In our work, we showcase two distinct approaches to obtain the point prompts. The first method relies on manual input, where users place positive and negative points on an interactive image window. The second approach automates the process by utilizing a keypoint detection model, which identifies precise points to be used as positive point prompts to SAM 2. This is demonstrated on the task of stem diameter estimation, where a keypoint detection model was trained to place two points on each of the foreground stems. Details on the automated approach are provided in Section \ref{sec:method-KPD}.

\subsection{Binary Mask Processing}
\label{sec:mask_processing}

Currently, precise control over the output masks generated by Segment Anything is limited, resulting in masks with imperfections that complicate the downstream task of skeleton construction (see Figure \ref{fig:framework_overview}). Typical issues include disconnected patches, holes within the binary mask, and rough contours, which lead to the generation of unnecessary branches during skeletonization. To address these issues, a minimal mask processing step is applied to refine the mask, enabling a smoother skeletonization process while preserving the shape of the original binary mask.

The binary mask processing pipeline begins by removing small, isolated objects from the initial mask that are smaller than a minimum size threshold. Next, connected component analysis is then performed to identify and retain the largest connected region, assumed to be the primary object of interest, while discarding smaller, irrelevant components.

Following connected component analysis, the mask undergoes a morphological opening operation followed by morphological closing. Both operations involve the fundamental steps of erosion and dilation, wherein a kernel of ones slides over the binary image. During erosion, a pixel is set to one only if all pixels within the kernel are one; otherwise, it is set to zero. Dilation operates in reverse, where a pixel is set to one if at least one pixel in the kernel is one. The opening operation, comprising erosion followed by dilation, has a smoothing effect on the object boundary by removing small protrusions. In contrast, the closing operation, which consists of dilation followed by erosion, enhances structural integrity by filling small internal gaps within the object.

Lastly, contour detection is then performed to delineate the outer boundary of the remaining, largest component. This boundary is approximated to a polygonal shape using the Ramer–Douglas–Peucker algorithm. The resulting polygonal representation is the filled, producing a clean, refined binary mask, that facilitates subsequent steps. 

\subsection{Skeleton Construction}\label{sec:method-skeleton}
In general context, skeletons are single-pixel-wide representations of a binary object that capture its essential topology \cite{blum1967transformation}. In our work, we aim to generate an ordered array of skeleton coordinates, where the first and last indices point to the bottommost and topmost skeleton pixel respectively, and adjacent indices represent connected skeleton pixels. This ordered structure encodes local slope information, which is necessary for identifying perpendicular line segments at any given skeleton pixel and enables accurate dimensional measurements. The process of deriving these ordered skeleton coordinates is referred to as skeleton construction.

Two approaches are proposed for skeleton construction based on the object's geometry. For rod-like geometries, the skeleton is obtained using standard skeletonization techniques, followed by pruning, augmentation, and reordering to ensure a structured and ordered representation. For regular objects with general geometries, Principal Component Analysis (PCA) is used to identify the object’s closest axis of symmetry, which guides the skeleton construction process. The distinction between these approaches arises from the inherent differences in geometry. Often times, the local slopes of general common shapes do not vary, and thus the slope can be captured with a single straight line. Also, general geometries result in diverse branching patterns, making it challenging to formulate a universal pruning method across all geometries.

For rod-like geometries, once the initial skeleton is obtained using a medial axis transform (MAT), the first step is to identify endpoints and intersection points. This is accomplished using the method described in \cite{choi2023mbest}. First, we compute the connectivity, $c_{ij}$, for each skeleton pixel, $s_{ij}$, by convolving a one-pixel border-padded version of the skeleton with the following kernel:
\begin{equation*}
    k = \begin{bmatrix}
        1 & 1 & 1 \\
        1 & 10 & 1 \\
        1 & 1 & 1 
        \end{bmatrix}
\end{equation*}
A pixel $s_{ij}$ is classified as an endpoint if $ c_{ij} = 11 $, and it is classified as an intersection point if $c_{ij} > 12$, which defines the sets:
\begin{equation*}
\mathbf{E} = \{ s_{ij} \mid c_{ij} = 11 \} \quad \text{and} \quad \mathbf{I} = \{ s_{ij} \mid c_{ij} > 12 \},
\end{equation*}
where $\mathbf{E}$ and $\mathbf{I}$ represent the set of endpoints and intersections respectively. However, defining the intersections based on connectivity alone cannot handle cases where adjacent intersection points are redundantly identified. To address this, clustering is performed on all identified intersections, ensuring that only one intersection pixel is preserved per cluster.

Following the identification of endpoints and intersection points, short branches—artifacts arising from rough contours that do not contribute to the topology—are detected and removed. This is achieved by iterating from each endpoint and determining its distance to the nearest intersection point or endpoint by propagating through neighboring skeleton pixels. A threshold value, $d$, set as the maximum diameter obtained from MAT, is used to classify branches. Branches with lengths shorter than $d$ is deemed a ``short" branch and is pruned. Conversely, branches exceeding this threshold are considered ``valid" and are retained.

Following the pruning step, the endpoints are re-evaluated on the pruned skeleton. A path is identified between endpoint candidates such that the vertical separation between them is maximized. This task is formulated as a minimum cost path problem, where the cost array is initialized such that skeleton pixels are assigned finite values while non-skeleton pixels are assigned infinity. By solving the minimum cost path problem with specified start and end points, the skeleton pixels along the optimal path can be identified.

Finally, the skeleton is augmented by extending additional segments from each endpoint. Extensions follow the local slope at each endpoint and are constrained to remain within the bounds of the binary mask. This augmentation ensures the presence of perpendicular line segments throughout the object's entire length. The augmented skeleton is reordered such that the first index corresponds to the bottommost skeleton pixel and the last index corresponds to the topmost pixel.

For general geometries, the skeleton is constructed using the symmetry axis identified through PCA. It produces two orthogonal principal axes: one corresponding to the direction of greatest variance and the other, its orthogonal axis. These axes are candidate skeleton axes, and the skeleton axis that is closer to the object's axis of symmetry is selected by comparing the dissimilarity between the original mask and its reflected masks derived from each candidate axis. Each axis bisects the original mask, $M_p$, into two parts. These parts are combined with their respective reflections about the candidate skeleton axis, and the dissimilarity scores are calculated. The dissimilarity score reflects the total number of different pixels of two binary masks, and the axis with the lower score is chosen as the skeleton axis:

\begin{align*}
    DS_1 &= \sum |M_p - M_r^{11}| + \sum|M_p - M_r^{12}| \\
    DS_2 &= \sum |M_p - M_r^{21}| + \sum |M_p - M_r^{22}|,
\end{align*}
where $M_r^{ij}$ denotes the reflected masks for each principal axis. For objects with multiple axes of symmetry, where the ratio of the minimum to the maximum dissimilarity scores is greater than 0.9, the first principal axis is automatically selected to ensure that perpendicular line segments are drawn along the object's longest dimension.

Once the skeleton axis is identified, skeleton construction begins from the object's centroid, propagating in both positive and negative directions along the axis. The skeleton is augmented iteratively while remaining within the bounds of the binary mask. Finally, the skeleton pixels are reordered from the bottommost to the topmost, which results in an augmented skeleton just like the rod-like geometry case.

\subsection{Simultaneous Line Segment and Depth Identification}
\label{sec:measurement_point_id}
All three-dimensional measurements provided by our framework rely on identifying line segments perpendicular to the object's skeleton, with their endpoints lying on the object's contour. Then, these endpoints, or their midpoints, are used as pixels of interest for transforming into 3D points. This section presents a method for simultaneously identifying these endpoints and retrieving their depth values.

Given an ordered list of skeleton pixels $\mathbf{S} = (s_1, s_2, \dots, s_n) $, where $ s_1 $ and $ s_n $ are the bottommost and topmost skeleton pixels, $\mathbf{S}'$ denotes the subset of sampled skeleton pixels at which perpendicular line segments are obtained. The first parameter $p_1$ is the sampling stride that determines the interval between sampled skeleton pixels. For each $ s_k' \in \mathbf{S}'$, its line segment is represented as the x, y and depth of each of its endpoints: $(x_{k1}, y_{k1}, d_{k1}, x_{k2}, y_{k2}, d_{k2})$.

To obtain this representation, the first step involves calculating the local slope at each selected skeleton pixel using the central difference method. For each skeleton pixel $ s_k' = (x_k, y_k) \in \mathbf{S}' $, its local slope $ \theta(s_k') $ is computed as:
\begin{equation*}
    \theta(s_k') = \arctan \left(\frac{y_{k+p_2} - y_{k-p_2}}{x_{k+p_2} - x_{k-p_2}}\right),
\end{equation*}
where $p_2$ specifies the offset in pixels on either side of 
$ s_k'$. For skeleton pixels near the endpoints, where such offsets may not exist, the calculation is adjusted to avoid out-of-bounds errors.


For estimating the line segment coordinates, we search along the perpendicular direction to the slope. Starting from the skeleton pixel, we search outwards until the mask boundary is reached. Then the median depth is selected for each endpoint of the line segment. Algorithm \ref{alg:dual_measurement_depth_simple} summarizes the steps for simultaneous line segment and depth identification. This iterative depth retrieval is necessary because the endpoints often lie on the object's contour, where depth data is frequently missing.  

\begin{algorithm}
\label{alg:dual_measurement_depth_simple}
\caption{Simultaneous Depth Estimation and Line Segment Identification}
\SetAlgoLined
\LinesNumbered
\DontPrintSemicolon
\KwIn{Sampled skeleton pixels $\mathbf{S}'$, Slopes $\boldsymbol{\Theta}$, Binary mask $\texttt{Mask}$, Depth map $\texttt{DepthMap}$}
\KwOut{Line segment coordinates $\mathbf{L} = \{(x_{k1}, y_{k1}, x_{k2}, y_{k2})\}$, Depths $\mathbf{D} = \{(d_{k1}, d_{k2})\}$}
\SetKwFunction{GetDepth}{GetDepth}
\SetKwFunction{Median}{Median}
\SetKwProg{Fn}{Func}{:}{}

\Fn{LineDepth($\mathbf{S}', \boldsymbol{\Theta}, \texttt{Mask}, \texttt{DepthMap}$)}{
    $\mathbf{L} \gets [ \ ], \mathbf{D} \gets [ \ ]$ \;

    \For{$s_k' = (x_k, y_k) \in \mathbf{S}'$ with slope $\theta_k \in \boldsymbol{\Theta}$}{
        $dx \gets -\sin(\theta_k), \ dy \gets \cos(\theta_k)$\;
        $(x_1, y_1) \gets (x_k, y_k)$, $(x_2, y_2) \gets (x_k, y_k)$\;
        $d_0 \gets \GetDepth(\texttt{DepthMap}, x_k, y_k)$\;
        Depth arrays: $\texttt{Depth1} \gets [\ ]$, $\texttt{Depth2} \gets [\ ]$\;

        \For{\textup{direction} $\in \{-1, +1\}$}{
            $(x, y) \gets (x_k, y_k)$\;
            \While{$(x, y)$ is within $\texttt{Mask}$}{
                $x \gets x + \textup{direction} \cdot dx$\;
                $y \gets y + \textup{direction} \cdot dy$\;
                $d \gets \GetDepth(\texttt{DepthMap}, x, y)$\;
                \If{$d$ \textup{exists and} $|d - d_0| < 0.1d_0$}{
                    \If{\textup{direction} == -1}{
                        Append $d$ to $\texttt{Depth1}$\;
                    }
                    \Else{
                        Append $d$ to $\texttt{Depth2}$\;
                    }
                }
                \Else{
                    \textbf{break}\;
                }
            }
            \If{\textup{direction} == -1}{
                $(x_1, y_1) \gets (x, y)$\;
            }
            \Else{
                $(x_2, y_2) \gets (x, y)$\;
            }
        }

        $d_1 \gets \Median(\texttt{Depth1})$\;
        $d_2 \gets \Median(\texttt{Depth2})$\;

        Append $(x_1, y_1, x_2, y_2)$ to $\mathbf{L}$\;
        Append $(d_1, d_2)$ to $\mathbf{D}$\;
    }
    \Return{$\mathbf{L}, \mathbf{D}$}\;
}
\end{algorithm}

\subsection{2D-3D Transform and Measurement}\label{sec:triangulation}

The 2D coordinates are projected into the 3D space by the traditional 2D-3D view transform using the intrinsics and distortion coefficients for the RGB and depth image.

The diameter obtained from the $k$th line segment, $D_k$ is the Euclidean distance between the endpoints in 3D of each valid line segment, given as $D_k = \| \textbf{X}_{k1} - \textbf{X}_{k2}\|$, where $\textbf{X}_{k1}$ and $\textbf{X}_{k2}$ are the 3D coordinates of the endpoints $k1$ and $k2$, respectively. The length of the object, $L$, is computed as the sum of the Euclidean distances between the 3D midpoints of consecutive line segments $l_k$, calculated using the average depth of its endpoints:
\begin{equation*}
    L = \sum_{k=1}^{n' - 1} l_k = \sum_{k=1}^{n' - 1}\| \textbf{X}_{k+1}^{\textrm{mid}} - \textbf{X}_{k}^{\textrm{mid}}\|,
\end{equation*}
Here $n'$ is the total number of line segments. Lastly, the volume of the object, $V$, is estimated by accumulating the volumes of partial cones formed by two consecutive line segments:
\begin{equation*}
    V = \sum_{k=1}^{n'-1}  \frac{1}{3}\pi l_k\left( \frac{D_k}{2}^2 + \frac{D_k D_{k+1}}{4} + \frac{D_{k+1}}{2}^2\right).
\end{equation*}

\begin{figure*}[!t]
    \centering
     \includegraphics[width=\textwidth]{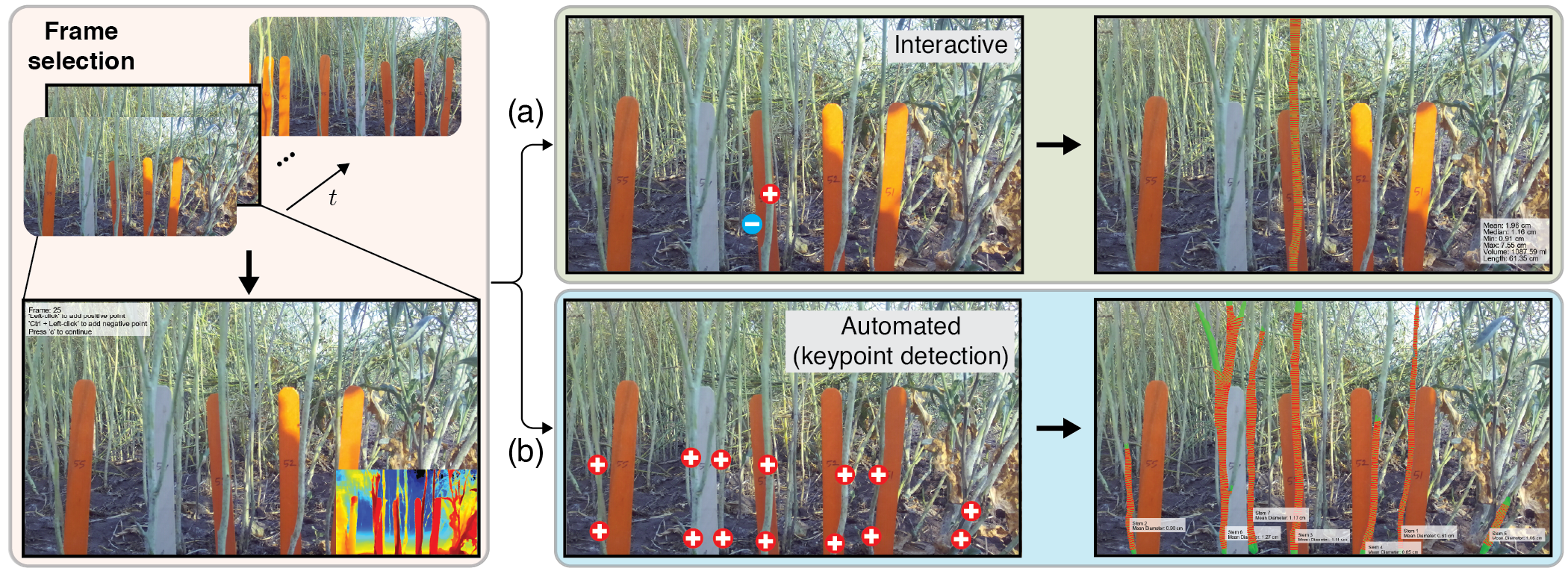}
    \caption{Demonstration of Measure Anything on Canola stems using the interactive, automated method. (a) Interactive method requires any number of positive / negative point prompts. (b) A trained keypoint detection model detects all foreground stems, whose outputs are used as positive point prompts.}
    \label{fig:results_stem_diameter}
\end{figure*}

\subsection{Keypoint Detection for Automating Prompts}\label{sec:method-KPD}


To enable fully automated point prompt generation for Measure Anything, we integrated a keypoint detection model. This approach was applied to large-scale diameter measurements on Canola stems, leveraging the efficiency of automated point generation provided by the keypoint detection model. For keypoint detection, we utilized the state-of-the-art YOLOv8 model. After testing various configurations with different numbers of keypoints per stem (1, 2, 3), our findings show that two keypoints per stem are sufficient in most cases for accurate segmentation and result in higher mean Average Precision (mAP) during training. The model was trained to focus exclusively on stems in the front rows of the image by annotating only these stems in the training dataset. A total of 320 carefully annotated images were used for training and testing, requiring approximately 50 hours of annotation effort, with each image taking roughly 8 minutes to annotate.

For crops with simpler stem structures, such as corn and wheat, or other applications involving regular-shaped objects, the automation task can be efficiently handled using bounding box object detection \cite{liu2023grounding}. Since Measure Anything is built on the Segment Anything framework, it supports both bounding box and point-based prompts, offering versatility and scalability for various downstream applications.

\begin{figure}[]
    \centering
     \includegraphics[width=\linewidth]{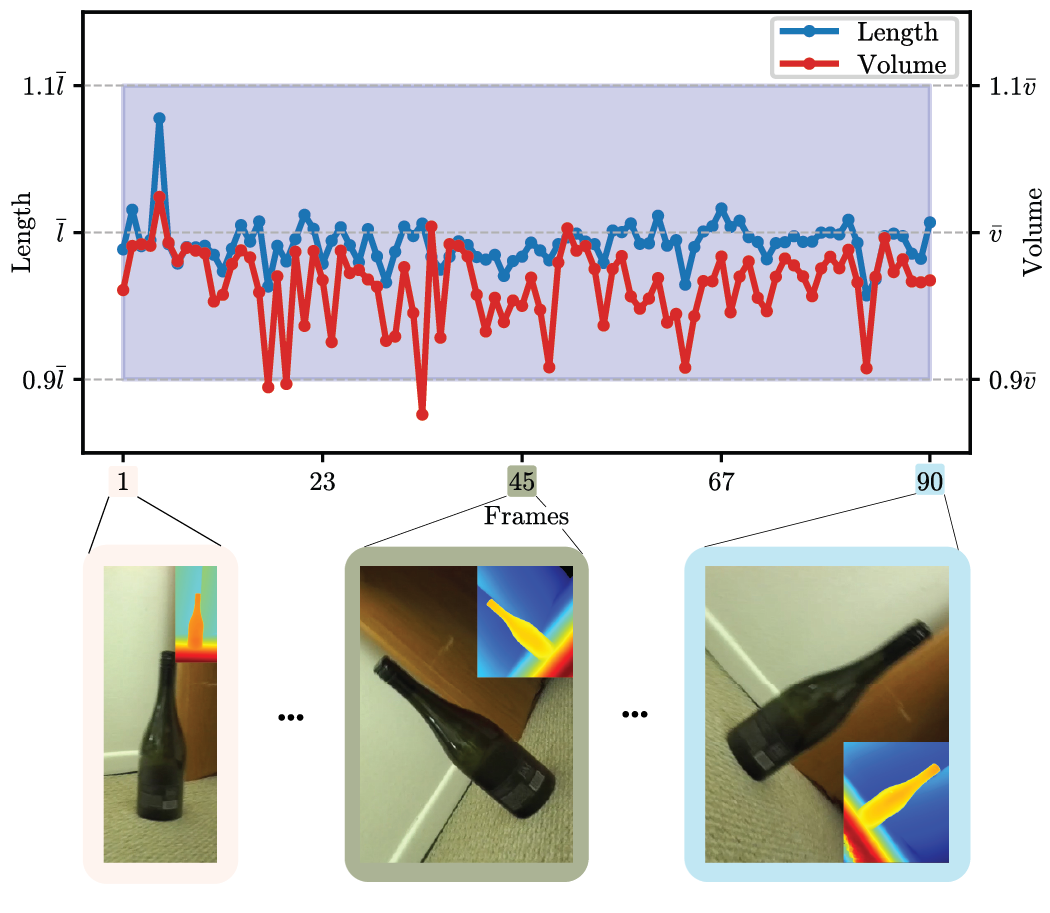}
    \caption{Variation in length and volume measurements of the object observed from different camera positions/  }
    \label{fig:results_variability}
\end{figure}

\subsection{Geometric Module for Robotic Grasping}

This study demonstrates the application of Measure Anything as a geometric module for robotic grasping tasks. The framework is evaluated using the Clubs dataset \cite{novkovic2019clubs}, which features cluttered images of various common objects inside a box, posing significant challenges for perception and manipulation tasks.
The primary roles of Measure Anything in this context are: (a) isolating an object from a cluttered environment and (b) extracting its geometric features. Segmenting and isolating objects in cluttered scenes is particularly critical when complete 3D point clouds are unavailable, as well as for sim-to-real transfer in robotic grasping models \cite{kasaei2023mvgrasp}. Extracting dimensions such as diameter profiles along the major principal axis, length, and volume provides valuable geometric priors for stability models used to identify optimal grasping points. Although missing depth information near object edges can pose challenges, Measure Anything can effectively obtain diameter measurements through the method described in Algorithm \ref{alg:dual_measurement_depth_simple}.

Within this work, the extracted geometric features are used in a simplified stability model to identify the top \textit{k} optimal grasping pairs of coordinates for a parallel gripper. The stability model incorporates three weighted factors: (i) the perpendicular distance from the center of gravity (CoG) \cite{fang2023anygrasp}, (ii) the diameter between the grasp coordinates, and (iii) Bicchi's criterion \cite{bicchi1995closure} for form closure. The equation of the model can be written as:
\begin{equation}\label{eq:stability}
    S_n = \underbrace{w_1 \cdot (1 - \hat{d_n}) + w_2 \cdot (1 - \hat{p_n})}_{\text{Force Closure}} + \underbrace{w_3 \cdot \text{cond}_n}_{\text{Form Closure}}
\end{equation}
where $S_n$ represents the stability score of $n$th pair of coordinates, $\hat{d_n}$ is the normalized diameter of the grip, $\hat{p_n}$ is the normalized perpendicular distance from the grasping line-segment to the CoG, and ``cond" is the condition if the pair of coordinates lie at a minima of a concave surface. Although the stability model covers both force closure and form closure components, we anticipate that state-of-the-art deep learning-based stability models \cite{fang2023anygrasp}, augmented with geometric priors from Measure Anything, would perform better across a broader spectrum of objects.

\section{RESULTS}\label{sec:results}

\begin{figure}[]
    \centering
     \includegraphics[width=\linewidth]{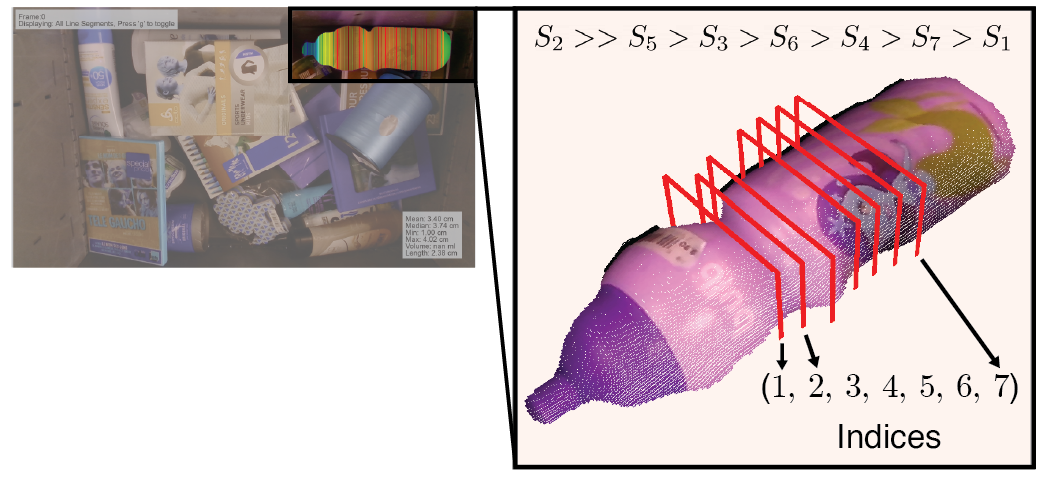}
    \caption{Diameter analysis using Measure Anything for identifying optimal grasp points of a standard object. }
    \label{fig:results_grasping}
\end{figure}

Figure \ref{fig:results_stem_diameter} illustrates the two demo versions discussed earlier in Sections \ref{sec:segmentation} and \ref{sec:method-KPD}. After selecting a frame from a video sequence captured by a stereo depth camera, Figure \ref{fig:results_stem_diameter}-(a) demonstrates the interactive method, where users can place any number of positive or negative point prompts via an interactive interface. Figure \ref{fig:results_stem_diameter}-(b) showcases the automated approach, which uses results from the keypoint detection model to generate point prompts. The keypoint detection model (YOLO v8), trained on a dataset of 256 images, achieved an average mAP-95 score of 83.5\%. A key advantage of the automated approach is the selective intervention feature, allowing users to correct keypoint labels when segmentation results are inaccurate. This correction is reinforced by adding additional positive or negative prompts.

Figure \ref{fig:results_variability} explores the variability of length and volume measurements obtained using the proposed framework across different camera positions. The analyzed video consists of approximately 90 frames of a wine bottle captured side-on, with variations in roll angles and distances from the object. Length and volume measurements across all frames are plotted and compared to the ground truth values, manually measured as 29.845 cm and 943 ml, respectively. Results show that most measurements remain consistent, varying within $\pm$ 10 \% of the ground truth values. It is noted that the quality of depth data is critical for accurate measurements, with outliers primarily arising from slightly inaccurate depth maps caused by blurry images resulting from camera motion.
 
Robotic grasp selection is demonstrated on an image from the Clubs dataset (see Figure \ref{fig:results_grasping}). The Measure Anything framework isolates the object of interest from a cluttered environment, processes the binary mask, and performs skeleton construction as well as line segment and depth identification, as detailed in previous sections. Stability scores for each line segment are then computed using Equation \ref{eq:stability}, and the top seven line segments with the highest scores are visualized in the figure. Notably, the line segment at the concave surface achieves the highest stability score, followed by the segments closest to the center of gravity (CoG), aligning well with intuitive expectations.

\section{CONCLUSION}
This work introduced a vision-based dimensional measurement framework that leverages the Segment Anything model to accurately generate object masks. The proposed pipeline demonstrated the ability to obtain precise diameter, length, and volume measurements across a variety of objects, including those with rod-like and general geometries. By utilizing continuous diameter measurements at different locations along the object, the framework was further shown to be applicable to additional tasks, such as identifying ideal gripping points for robotic manipulation.

Future work will focus on addressing the limitations of the current framework. For instance, segmentation models are susceptible to inaccuracies caused by occlusions from overlapping objects, which can result in erroneous dimensional measurements. We aim to enhance the robustness of our approach by incorporating shape priors to handle such occlusions. Additionally, the current pipeline assumes objects with circular cross-sections; future iterations will integrate intelligent models capable of identifying object types, paving the way for strategies tailored to objects with non-circular cross-sections. We also plan to integrate Measure Anything with a deep learning grasping model to compare its performance against the state-of-the-art. Finally, we plan to explore the integration of language models to enable diverse tasks, such as obtaining dimensional measurements of specific objects within complex scenes.

\section*{ACKNOWLEDGMENT}
The authors would like to thank Mohammad Jony, Md Fahad Hassan, Cristhian Perdigon, Dr. Md Mukhlesur Rahman, Dr. Paulo Flores from North Dakota State University for their efforts with data collection from the greenhouse and fields of North Dakota.

\bibliography{refs}

\begin{thebibliography}{10}

\bibitem{kirillov2023segment}
Alexander Kirillov, Eric Mintun, Nikhila Ravi, Hanzi Mao, Chloe Rolland, Laura Gustafson, Tete Xiao, Spencer Whitehead, Alexander~C Berg, Wan-Yen Lo, et~al.
\newblock Segment anything.
\newblock {\em arXiv preprint arXiv:2304.02643}, 2023.

\bibitem{ravi2024sam}
Nikhila Ravi, Valentin Gabeur, Yuan-Ting Hu, Ronghang Hu, Chaitanya Ryali, Tengyu Ma, Haitham Khedr, Roman R{\"a}dle, Chloe Rolland, Laura Gustafson, et~al.
\newblock Sam 2: Segment anything in images and videos.
\newblock {\em arXiv preprint arXiv:2408.00714}, 2024.

\bibitem{de2015stem}
Tom De~Swaef, Veerle De~Schepper, Maurits~W Vandegehuchte, and Kathy Steppe.
\newblock Stem diameter variations as a versatile research tool in ecophysiology.
\newblock {\em Tree Physiology}, 35(10):1047--1061, 2015.

\bibitem{kim2024p}
Kitae Kim, Aarya Deb, and David~J Cappelleri.
\newblock P-agslam: In-row and under-canopy slam for agricultural monitoring in cornfields.
\newblock {\em IEEE Robotics and Automation Letters}, 2024.

\bibitem{zhou2021rgb}
Tao Zhou, Deng-Ping Fan, Ming-Ming Cheng, Jianbing Shen, and Ling Shao.
\newblock Rgb-d salient object detection: A survey.
\newblock {\em Computational Visual Media}, 7:37--69, 2021.

\bibitem{wang2021development}
Rongxuan Wang, Andrew~C Law, David Garcia, Shuo Yang, and Zhenyu Kong.
\newblock Development of structured light 3d-scanner with high spatial resolution and its applications for additive manufacturing quality assurance.
\newblock {\em The International Journal of Advanced Manufacturing Technology}, 117:845--862, 2021.

\bibitem{shang2022measurement}
Hang Shang, Changying Liu, and Ruijian Wang.
\newblock Measurement methods of 3d shape of large-scale complex surfaces based on computer vision: A review.
\newblock {\em Measurement}, 197:111302, 2022.

\bibitem{kang2020fruit}
Hanwen Kang and Chao Chen.
\newblock Fruit detection, segmentation and 3d visualisation of environments in apple orchards.
\newblock {\em Computers and Electronics in Agriculture}, 171:105302, 2020.

\bibitem{mirbod2023tree}
Omeed Mirbod, Daeun Choi, Paul~H Heinemann, Richard~P Marini, and Long He.
\newblock On-tree apple fruit size estimation using stereo vision with deep learning-based occlusion handling.
\newblock {\em Biosystems Engineering}, 226:27--42, 2023.

\bibitem{vit2019length}
Adar Vit, Guy Shani, and Aharon Bar-Hillel.
\newblock Length phenotyping with interest point detection.
\newblock In {\em Proceedings of the IEEE/CVF Conference on Computer Vision and Pattern Recognition Workshops}, pages 0--0, 2019.

\bibitem{xiang2021measuring}
Lirong Xiang, Lie Tang, Jingyao Gai, and Le~Wang.
\newblock Measuring stem diameter of sorghum plants in the field using a high-throughput stereo vision system.
\newblock {\em Transactions of the ASABE}, 64(6):1999--2010, 2021.

\bibitem{he2017mask}
Kaiming He, Georgia Gkioxari, Piotr Doll{\'a}r, and Ross Girshick.
\newblock Mask r-cnn.
\newblock In {\em Proceedings of the IEEE international conference on computer vision}, pages 2961--2969, 2017.

\bibitem{hirschmuller2007stereo}
Heiko Hirschmuller.
\newblock Stereo processing by semiglobal matching and mutual information.
\newblock {\em IEEE Transactions on pattern analysis and machine intelligence}, 30(2):328--341, 2007.

\bibitem{ma2024computer}
Weihong Ma, Yi~Sun, Xiangyu Qi, Xianglong Xue, Kaixuan Chang, Zhankang Xu, Mingyu Li, Rong Wang, Rui Meng, and Qifeng Li.
\newblock Computer-vision-based sensing technologies for livestock body dimension measurement: A survey.
\newblock {\em Sensors}, 24(5):1504, 2024.

\bibitem{fernandes2020image}
Arthur Francisco~Ara{\'u}jo Fernandes, Jo{\~a}o Ricardo~Rebou{\c{c}}as D{\'o}rea, and Guilherme Jord{\~a}o de~Magalh{\~a}es Rosa.
\newblock Image analysis and computer vision applications in animal sciences: an overview.
\newblock {\em Frontiers in Veterinary Science}, 7:551269, 2020.

\bibitem{bicchi2000robotic}
Antonio Bicchi and Vijay Kumar.
\newblock Robotic grasping and contact: A review.
\newblock In {\em Proceedings 2000 ICRA. Millennium conference. IEEE international conference on robotics and automation. Symposia proceedings (Cat. No. 00CH37065)}, volume~1, pages 348--353. IEEE, 2000.

\bibitem{marwan2021comprehensive}
Qaid~Mohammed Marwan, Shing~Chyi Chua, and Lee~Chung Kwek.
\newblock Comprehensive review on reaching and grasping of objects in robotics.
\newblock {\em Robotica}, 39(10):1849--1882, 2021.

\bibitem{zhang2020state}
Baohua Zhang, Yuanxin Xie, Jun Zhou, Kai Wang, and Zhen Zhang.
\newblock State-of-the-art robotic grippers, grasping and control strategies, as well as their applications in agricultural robots: A review.
\newblock {\em Computers and Electronics in Agriculture}, 177:105694, 2020.

\bibitem{du2021vision}
Guoguang Du, Kai Wang, Shiguo Lian, and Kaiyong Zhao.
\newblock Vision-based robotic grasping from object localization, object pose estimation to grasp estimation for parallel grippers: a review.
\newblock {\em Artificial Intelligence Review}, 54(3):1677--1734, 2021.

\bibitem{cheng2022vision}
Hu~Cheng, Yingying Wang, and Max Q-H Meng.
\newblock A vision-based robot grasping system.
\newblock {\em IEEE Sensors Journal}, 22(10):9610--9620, 2022.

\bibitem{fang2020graspnet}
Hao-Shu Fang, Chenxi Wang, Minghao Gou, and Cewu Lu.
\newblock Graspnet-1billion: A large-scale benchmark for general object grasping.
\newblock In {\em Proceedings of the IEEE/CVF conference on computer vision and pattern recognition}, pages 11444--11453, 2020.

\bibitem{wang2021graspness}
Chenxi Wang, Hao-Shu Fang, Minghao Gou, Hongjie Fang, Jin Gao, and Cewu Lu.
\newblock Graspness discovery in clutters for fast and accurate grasp detection.
\newblock In {\em Proceedings of the IEEE/CVF International Conference on Computer Vision}, pages 15964--15973, 2021.

\bibitem{fang2023anygrasp}
Hao-Shu Fang, Chenxi Wang, Hongjie Fang, Minghao Gou, Jirong Liu, Hengxu Yan, Wenhai Liu, Yichen Xie, and Cewu Lu.
\newblock Anygrasp: Robust and efficient grasp perception in spatial and temporal domains.
\newblock {\em IEEE Transactions on Robotics}, 2023.

\bibitem{minaee2021image}
Shervin Minaee, Yuri Boykov, Fatih Porikli, Antonio Plaza, Nasser Kehtarnavaz, and Demetri Terzopoulos.
\newblock Image segmentation using deep learning: A survey.
\newblock {\em IEEE transactions on pattern analysis and machine intelligence}, 44(7):3523--3542, 2021.

\bibitem{long2015fully}
Jonathan Long, Evan Shelhamer, and Trevor Darrell.
\newblock Fully convolutional networks for semantic segmentation.
\newblock In {\em Proceedings of the IEEE conference on computer vision and pattern recognition}, pages 3431--3440, 2015.

\bibitem{noh2015learning}
Hyeonwoo Noh, Seunghoon Hong, and Bohyung Han.
\newblock Learning deconvolution network for semantic segmentation.
\newblock In {\em Proceedings of the IEEE international conference on computer vision}, pages 1520--1528, 2015.

\bibitem{badrinarayanan2017segnet}
Vijay Badrinarayanan, Alex Kendall, and Roberto Cipolla.
\newblock Segnet: A deep convolutional encoder-decoder architecture for image segmentation.
\newblock {\em IEEE transactions on pattern analysis and machine intelligence}, 39(12):2481--2495, 2017.

\bibitem{ronneberger2015u}
Olaf Ronneberger, Philipp Fischer, and Thomas Brox.
\newblock U-net: Convolutional networks for biomedical image segmentation.
\newblock In {\em Medical image computing and computer-assisted intervention--MICCAI 2015: 18th international conference, Munich, Germany, October 5-9, 2015, proceedings, part III 18}, pages 234--241. Springer, 2015.

\bibitem{cciccek20163d}
{\"O}zg{\"u}n {\c{C}}i{\c{c}}ek, Ahmed Abdulkadir, Soeren~S Lienkamp, Thomas Brox, and Olaf Ronneberger.
\newblock 3d u-net: learning dense volumetric segmentation from sparse annotation.
\newblock In {\em Medical Image Computing and Computer-Assisted Intervention--MICCAI 2016: 19th International Conference, Athens, Greece, October 17-21, 2016, Proceedings, Part II 19}, pages 424--432. Springer, 2016.

\bibitem{dosovitskiy2020image}
Alexey Dosovitskiy.
\newblock An image is worth 16x16 words: Transformers for image recognition at scale.
\newblock {\em arXiv preprint arXiv:2010.11929}, 2020.

\bibitem{carion2020end}
Nicolas Carion, Francisco Massa, Gabriel Synnaeve, Nicolas Usunier, Alexander Kirillov, and Sergey Zagoruyko.
\newblock End-to-end object detection with transformers.
\newblock In {\em European conference on computer vision}, pages 213--229. Springer, 2020.

\bibitem{li2023mask}
Feng Li, Hao Zhang, Huaizhe Xu, Shilong Liu, Lei Zhang, Lionel~M Ni, and Heung-Yeung Shum.
\newblock Mask dino: Towards a unified transformer-based framework for object detection and segmentation.
\newblock In {\em Proceedings of the IEEE/CVF Conference on Computer Vision and Pattern Recognition}, pages 3041--3050, 2023.

\bibitem{blum1967transformation}
Harry Blum.
\newblock A transformation for extracting new descriptions of shape.
\newblock {\em Models for the perception of speech and visual form}, pages 362--380, 1967.

\bibitem{choi2023mbest}
Andrew Choi, Dezhong Tong, Brian Park, Demetri Terzopoulos, Jungseock Joo, and Mohammad~Khalid Jawed.
\newblock mbest: Realtime deformable linear object detection through minimal bending energy skeleton pixel traversals.
\newblock {\em arXiv preprint arXiv:2302.09444}, 2023.

\bibitem{liu2023grounding}
Shilong Liu, Zhaoyang Zeng, Tianhe Ren, Feng Li, Hao Zhang, Jie Yang, Qing Jiang, Chunyuan Li, Jianwei Yang, Hang Su, et~al.
\newblock Grounding dino: Marrying dino with grounded pre-training for open-set object detection.
\newblock {\em arXiv preprint arXiv:2303.05499}, 2023.

\bibitem{novkovic2019clubs}
Tonci Novkovic, Fadri Furrer, Marko Panjek, Margarita Grinvald, Roland Siegwart, and Juan Nieto.
\newblock Clubs: An rgb-d dataset with cluttered box scenes containing household objects.
\newblock {\em The International Journal of Robotics Research}, 38(14):1538--1548, 2019.

\bibitem{kasaei2023mvgrasp}
Hamidreza Kasaei and Mohammadreza Kasaei.
\newblock Mvgrasp: Real-time multi-view 3d object grasping in highly cluttered environments.
\newblock {\em Robotics and Autonomous Systems}, 160:104313, 2023.

\bibitem{bicchi1995closure}
Antonio Bicchi.
\newblock On the closure properties of robotic grasping.
\newblock {\em The International Journal of Robotics Research}, 14(4):319--334, 1995.

\end{thebibliography}
\bibliographystyle{unsrt}

\addtolength{\textheight}{-12cm}   

\end{document}